\documentclass[10pt,twocolumn,letterpaper]{article}

\usepackage[pagenumbers]{cvpr} 
\newcommand{\todo}[1]{{\color{red}#1}}

\definecolor{cvprblue}{rgb}{0.21,0.49,0.74}
\usepackage[pagebackref,breaklinks,colorlinks,allcolors=cvprblue]{hyperref}

\def\paperID{13618} 
\def\confName{CVPR}
\def\confYear{2025}

\title{Joint Learning of Depth and Appearance for Portrait Image Animation}

\author{Xinya Ji$^{1,2}$
	\hspace{7mm}
	Gaspard Zoss$^{3}$
	\hspace{7mm}
	Prashanth Chandran$^{3}$
	\\
	Lingchen Yang$^{1}$
	\hspace{7mm}
	Xun Cao$^{2}$
	\hspace{7mm}
	Barbara Solenthaler$^{1}$
	\hspace{7mm}
	Derek Bradley$^{3}$
	\\
    $^{1}$ETH Z\"urich
	\hspace{7mm}
	$^{2}$Nanjing University
	\hspace{7mm}
	$^{3}$DisneyResearch\textbar Studios \\
	{{\tt\small xinya@smail.nju.edu.cn}}
    \hspace{7mm} {{\tt\small caoxun@nju.edu.cn}}
	\hspace{7mm} {\tt\small {\{lingchen.yang,solenthaler\}@inf.ethz.ch}} \\
	{\tt\small {\{prashanth.chandran,gaspard.zoss,derek.bradley\}@disneyresearch.com}}
}

\begin{document}
\maketitle

\newcommand{\figref}[1]{Fig.~\ref{#1}}
\newcommand{\tabref}[1]{Table~\ref{#1}}
\newcommand{\eqnref}[1]{Eq.~\ref{#1}}
\newcommand{\secref}[1]{Section~\ref{#1}}
\newcommand{\appref}[1]{Appendix~\ref{#1}}


\newcommand{\shortcite}[1]{\cite{#1}}


\definecolor{dbcolor}{RGB}{50,10,210}
\newcommand\db[1] {{\textcolor{dbcolor}{\em\textbf{DB}: #1}}}

\definecolor{pccolor}{RGB}{10,150,100}
\newcommand\pc[1] {{\textcolor{pccolor}{\em\textbf{PC}: #1}}}

\definecolor{xjcolor}{RGB}{210,10,210}
\newcommand\xj[1] {{\textcolor{xjcolor}{\em\textbf{XJ}: #1}}}

\definecolor{gzcolor}{RGB}{210,210,10}
\newcommand\gz[1] {{\textcolor{gzcolor}{\em\textbf{GZ}: #1}}}

\definecolor{lycolor}{RGB}{10,210,50}
\newcommand\ly[1] {{\textcolor{lycolor}{\em\textbf{LY}: #1}}}

\definecolor{bscolor}{RGB}{10,150,100}
\newcommand\bs[1] {{\textcolor{bscolor}{\em\textbf{BS}: #1}}}

\definecolor{delcolor}{RGB}{210,0,0}
\definecolor{addcolor}{RGB}{0,0,0}
\newcommand\del[1] {{\textcolor{delcolor}{#1}}}
\newcommand\add[1] {{\textcolor{addcolor}{#1}}}

\renewcommand\todo[1] {{\textcolor{red}{\em\textbf{TODO}: #1}}}
\newcommand\camready[1] {{\textcolor{blue}{#1}}}

\makeatletter
\newcommand\footnoteref[1]{\protected@xdef\@thefnmark{\ref{#1}}\@footnotemark}
\makeatother

\renewcommand{\thefootnote}{\fnsymbol{footnote}}

\clubpenalty=10000
\widowpenalty=10000
\displaywidowpenalty=10000

\begin{abstract}

2D portrait animation has experienced significant advancements in recent years. Much research has utilized the prior knowledge embedded in large generative diffusion models to enhance high-quality image manipulation.
 However, most methods only focus on generating RGB images as output, and the co-generation of consistent visual plus 3D output remains largely under-explored.  In our work, we propose to jointly learn the visual appearance and depth simultaneously in a diffusion-based portrait image generator. Our method embraces the end-to-end diffusion paradigm and introduces a new architecture suitable for learning this conditional joint distribution, consisting of a reference network and a channel-expanded diffusion backbone. Once trained, our framework can be efficiently adapted to various downstream applications, such as facial depth-to-image and image-to-depth generation, portrait relighting, and audio-driven talking head animation with consistent 3D output.

\end{abstract}    
\section{Introduction}
\label{sec:intro}

A popular application in the field of computer vision is to edit or animate portrait photos.  From an editing perspective, researchers have devised a host of algorithms for tasks like portrait relighting~\cite{rao2024lite2relight, pandey2021total, kim2024switchlight}, facial expression manipulation~\cite{drobyshev2022megaportraits, ma2024follow,siarohin2019first,wang2021one, ramesh2022hierarchical, saharia2022photorealistic} and appearance editing~\cite{shen2020interpreting, kim2022diffusionclip, cheng20243d, yue2023chatface} (\eg changing the hair color, adding accessories or re-aging the person).  From an animation perspective, common tasks include animation retargeting from a driving video of another individual~\cite{guo2024liveportrait, xie2024x, jiang2024mobileportrait, yang2024megactor}, or synthetic talking head generation driven from an audio input~\cite{peng2024synctalk,yu2024gaussiantalker,zhou2020makelttalk,zhang2023sadtalker, zhou2021pose,ji2021audio, jiang2024loopy, xu2023omniavatar, zhuang2024vlogger, zhang2022unsupervised, ma2023otavatar}.

To accomplish these tasks, modern approaches rely on deep generative models, like GANs~\cite{karras2019style, karras2020analyzing} and diffusion models~\cite{ho2020denoising, blattmann2023stable, peebles2023scalable}, that are trained on large datasets of face images or videos.  What we learned from early models like StyleGAN~\cite{sauer2022stylegan, karras2019style} is that image synthesis methods can produce stunning photo-realistic images of human faces, indistinguishable from reality.  Thus the idea of recent image editing and animation approaches is to start with a pre-trained generative model~\cite{blattmann2023stable, dhariwal2021diffusion, radford2021learning, peebles2023scalable} as the backbone and learn to condition the generation on other signals, like illumination or audio.  While such methods~\cite{xu2024magicanimate, hu2024animate, guo2023animatediff, guo2023sparsectrl} have proven to be very powerful in portrait image manipulation, one issue is that previous methods only learn to generate the appearance (\ie the RGB color) of the face, which can limit downstream applications.  In this work we will show that jointly learning both appearance and depth will allow for several expanded applications in the field of portrait image animation.

We propose a new architecture for jointly learning both the depth and the appearance of faces in a generative model, designed for portrait image manipulation.  The correlation between the appearance and depth channels is of critical importance, \ie the generated depth map must match the generated face image.  We accomplish this with a new diffusion-based portrait image generator built on top of Stable Diffusion~\cite{peebles2023scalable, rombach2022high}, but adapted to learn this joint distribution.  Similar to related work~\cite{hu2024animate,xu2024hallo, tian2024emo}, we employ a reference network designed to extract the identity of an RGB reference photo, which guides the image diffusion process.  We expand the traditional Stable Diffusion backbone to de-noise a 6-channel input image, which consists of separately-noised RGB and depth latent images.  The shared UNet in the diffusion step ensures good correlation between the appearance and depth outputs.  Finally we train the model on a combination of studio-captured facial images with ground truth 3D geometry obtained from a facial scanner, and also in-the-wild facial videos with approximate 3D reconstructed geometry.  As we will show, this combination allows our model to both learn accurate depth generation and also generalize to outdoor settings.

Once our model is trained it can be adapted for several applications.  In addition to unconditional sampling to achieve coupled RGB + depth images, we show applications of channel-wise inpainting.  Specifically, for a given image we can inpaint the depth channel, achieving facial depth estimation with our model.  Alternatively, for a given depth image, we can inpaint the RGB channels to obtain an artistic way to control face image generation using either a 3D morphable face model or the estimated depth from a separate image.  Given the joint generation of appearance and depth, we can further relight the face image in a post-process using the normals from the depth image.  Finally, we show that our model can be extended to the task of creating audio-driven talking head videos, with paired appearance and depth that is consistent during the performance.

Specifically, our contributions are:
\begin{enumerate}
\item A novel architecture for joint learning of depth and appearance of portrait images,
\item A new training scheme for learning paired image and depth maps from a combination of in-studio and in-the-wild facial data,
\item The demonstration of several applications in portrait manipulation including both image-to-depth and depth-to-image channel-wise inpainting, portrait relighting, and audio-driven talking head animation.
\end{enumerate}

\section{Related Works}
\label{sec:realted}

\noindent {\bf Diffusion Model for Character Animation.}
Diffusion-based generative models have shown remarkable capabilities in generative tasks, demonstrating diversity and adaptability across various multimedia domains. The development of large pre-trained models, such as Stable Diffusion~\cite{rombach2022high}, has spurred numerous applications leveraging its robust model priors. By extending the pre-trained model from 2D image generation to 3D video generation, researchers have explored tasks for animating human images. For example, AnimateDiff~\cite{guo2023animatediff} introduces a plug-and-play temporal module designed to adapt flexibly to different motion patterns without model-specific tuning.  In animating specific characters, DreamPose~\cite{karras2023dreampose} introduces a dual clip-image encoder for image encoding. Similarly, methods like Animate Anyone~\cite{hu2024animate}, MagicAnimate~\cite{xu2024magicanimate}, and Talk-Act~\cite{guan2024talk} resort to a ReferenceNet with symmetrical U-Net architecture to maintain appearance consistency. Intermediate representations like landmarks, skeletons, or segmentation maps are used as control signals in this process for fine-grained control. Our work builds upon the diffusion priors of Stable Diffusion, achieving video generation by integrating a motion module for improved temporal consistency. 

\noindent {\bf Diffusion Model for Geometric Estimation.}
Diffusion models trained on large image datasets for high-quality generation tasks have been proven to contain a rich understanding of the underlying scene structure. 
This capability has extended the diffusion model to 3D geometric estimation tasks, including depth estimation~\cite{bhat2021adabins, bhat2023zoedepth, guizilini2023towards, jafarian2021learning, yang2024depth}, normal estimation~\cite{wang2015designing, xiu2023econ}, and view synthesis~\cite{lu2024direct2, shi2023mvdream, long2024wonder3d, wang2021neus}. 
Recently, Marigold~\cite{ke2024repurposing} leverages the diffusion priors by fine-tuning large pre-trained diffusion models specifically for depth estimation. 
Wonder3D~\cite{long2024wonder3d} designs a cross-domain diffusion model with attention across different modalities for information exchange. 
Geowizard~\cite{fu2025geowizard} proposes to jointly estimate depth and normals and involve a scene distribution decoupler strategy to discern different scene layouts. Recently, Khirodkar \etal ~\cite{khirodkar2025sapiens} proposes Sapiens, a human-centric foundation model capable of pose estimation, body segmentation, depth and normals estimation.
Several approaches have been developed to jointly denoise cross-domain representations utilizing the prior of large pretrained model. JointNet~\cite{zhang2023jointnet} achieves joint generation by replicating the original network, enabling it to handle multiple geometric tasks within a unified framework. Additionally, HyperHuman~\cite{liu2023hyperhuman} proposes to learn the correlation between appearance and geometric structure by denosing the depth and surface-normal along with the RGB image. Similarly, we design a unified framework to jointly learn RGB image and 3D depth by expanding the diffusion unet and utilizing the priors of large pretrained generative model.

\noindent {\bf Audio-Driven Talking Head Generation.}
The goal of audio-driven talking head generation is to synthesize facial movements synchronized with the driven audio~\cite{chen2018lip, vougioukas2020realistic, chung2017you, suwajanakorn2017synthesizing, ji2021audio, gururani2022spacex, zhou2020makelttalk, zhong2023identity, zhang2023sadtalker, zhou2020makelttalk, zhou2021pose, ji2022eamm, wang2023progressive, ma2023dreamtalk}. 
Recently, with the development of the diffusion model, much progress has been made in this area, emphasizing high-fidelity animation with synchronized lip motion. For instance, EMO~\cite{tian2024emo} firstly presents an end-to-end framework capable of generating realistic facial animations leveraging the pre-trained stable diffusion model from an audio track. Similarly, VASA-1~\cite{xu2024vasa} and AniTalker~\cite{liu2024anitalker} train an audio-conditioned diffusion model on the motion latent space for human faces and enables real-time generation. Subsequent work like Aniportrait~\cite{wei2024aniportrait}, EchoMimic~\cite{chen2024echomimic}, and V-Express~\cite{wang2024v} expand the model for subtle dynamics and stability by involving control signals like landmarks and sophisticated loss functions. Hallo~\cite{xu2024hallo} and its updated version Hallo2~\cite{cui2024hallo2} then focus on enhanced lip motion, long-duration and high-resolution animation. Loopy~\cite{jiang2024loopy} further improves the motion quality by expanding the receptive field of the motion frames. Other works like CyberHost~\cite{lin2024cyberhost} propose to integrate hand motion, which enables video generation within the scope of the human body. However, none of these previous works consider jointly synthesizing 2D and 3D animation by utilizing the 3D priors in the pre-trained diffusion model. Instead, we build a joint-learning framework which proves can be extended to the audio-driven task and generate smooth animation.

\section{Method}
\label{sec:method}

\begin{figure}
    \includegraphics[width=1.0\linewidth]{{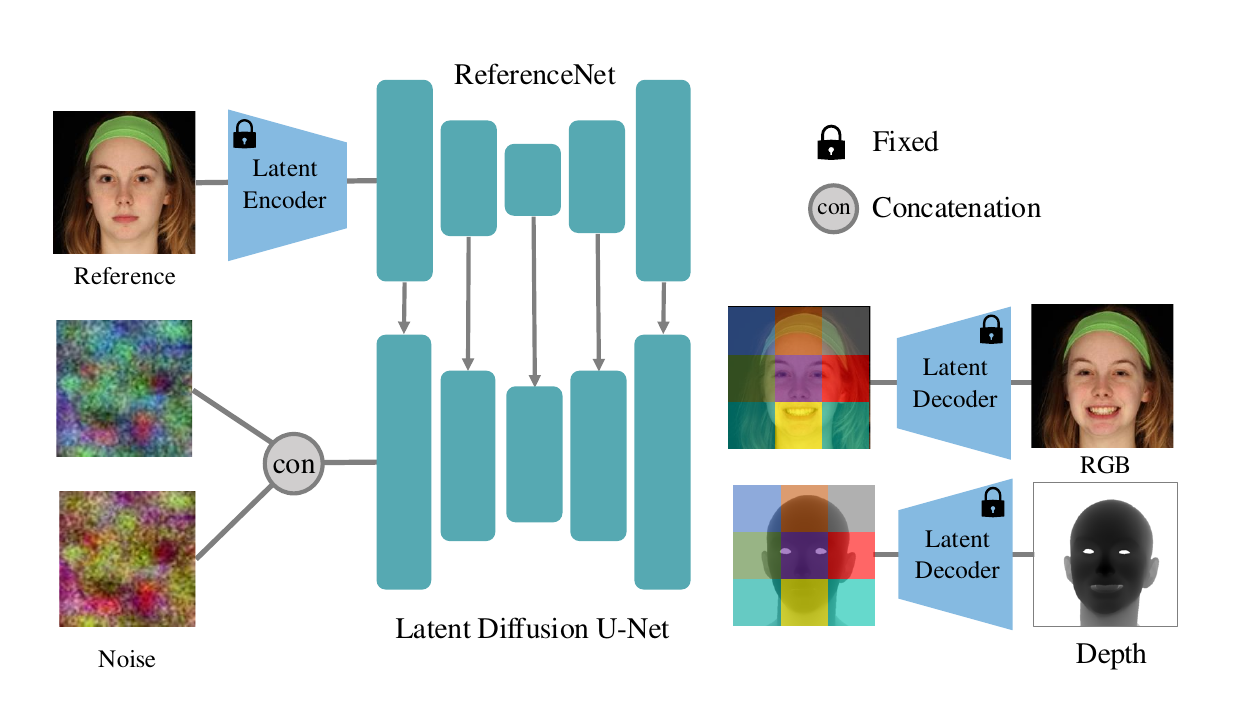}}
    \caption{The overview of the proposed pipeline. Given a reference image, our model jointly generates the appearance (RGB) and depth of the identity under various expressions and poses, by simply sampling random noise in the latent space.}
    \label{fig:pipeline}
\end{figure}

\subsection{Preliminaries}

\paragraph{Latent Diffusion Models.}
Diffusion models have set the new standard for generative models due to its ability to generate high-quality samples and perform a wide range of tasks with finetuning techniques.
The diffusion model is trained to generate images by iteratively adding noise to the image and then removing the noise level-by-level, so that the model learns to generate the image from the gaussian noise. Different from the diffusion models that directly work on the image space, the latent diffusion models perform diffusion in a latent space, providing computational compactness and scalability to higher resolution images. The latent space is obtained from a pretrained variational auto-encoder (VAE)~\cite{kingma2013auto}. 

For a given sampled image $\mathbf{x}$, the encoder $\mathcal{E}$ of the VAE encodes the image into this latent space, as $\mathbf{z} = \mathcal{E}(\mathbf{x})$. The forward pass of the diffusion process adds noise to the latent code $\mathbf{z}_0$ according to the uniformly sampled noise level $l$:
\begin{equation}
    \mathbf{z}_l = \sqrt{\bar{\alpha}_l} \mathbf{z}_0 + \sqrt{1 - \bar{\alpha}_l} \boldsymbol{\epsilon},
\end{equation}
where $\boldsymbol{\epsilon} \sim \mathcal{N}(0, \mathbf{I})$, $\bar{\alpha}_l$ is associated with the variance schedule of a diffusion process with $L$ noise levels so that $\mathbf{z}_L$ becomes a gaussian distribution. In the reverse process, the denoising network $\boldsymbol{\epsilon}_\theta(\cdot)$, parameterized with learnable parameters \(\theta\), gradually removes noise from $\mathbf{z}_l$ to get $\mathbf{z}_{l-1}$, so as to obtain the fully denoised $\mathbf{z}_{0}$. The decoder $\mathcal{D}$ of the VAE then decodes $\mathbf{z}_{0}$ to generate the image $\mathbf{x}$. During training, the parameters $\theta$ are updated by minimizing the following loss function:
\begin{equation}
    \mathcal{L}(\theta) = \mathbb{E}_{\boldsymbol{\epsilon} \sim \mathcal{N}(0, \mathbf{I}), l \sim \mathcal{U}(L)} \left\| \epsilon - \boldsymbol{\epsilon}_\theta(\mathbf{z}_l, l) \right\|^2 .
\end{equation}

Equipped with conditional information injected using cross-attention modules~\cite{vaswani2017attention}, the latent diffusion models can be extended to perform various tasks, such as text-to-image generation~\cite{rombach2022high} and image-to-image translation~\cite{zhang2023adding}. In this work, we propose to leverage a pretrained latent diffusion model and adapt it to perform the task of co-generation of depth and appearance for portrait image animation, conditioned on a reference image.

\subsection{Joint Learning of Depth and Appearance}
As demonstrated in \figref{fig:pipeline}, given a reference image $\mathbf{r}$ of the identity of interest, our task is to jointly generate the appearance (RGB) $\mathbf{x}$ and depth $\mathbf{d}$ of the subject under various expressions and poses. We model this as a conditional joint distribution in the latent diffusion U-Net~\cite{ronneberger2015u} model, as $p(\mathbf{z}^\mathbf{x}_0, \mathbf{z}_0^\mathbf{d} | \mathbf{r})$, where $\mathbf{z}^\mathbf{x}_0$ and $\mathbf{z}_0^\mathbf{d}$ are the latent features for the appearance and depth, respectively.
The final maps are generated by decoding the latent codes with the decoder of the VAE, as $\mathbf{x} = \mathcal{D}(\mathbf{z}^\mathbf{x}_0)$ and $\mathbf{d} = \mathcal{D}(\mathbf{z}^\mathbf{d}_0)$.

The reference image $\mathbf{r}$ is essential to generate consistent appearance and depth of the identity. In order to capture intricate details of the target, we use a reference network $\mathcal{R}$ to extract the identity features $\mathbf{z}^\mathbf{r}$ from the reference image $\mathbf{r}$, as $\mathbf{z}^\mathbf{r} = \mathcal{R}(\mathbf{r})$, which are then injected into the latent diffusion model using the spatial attention modules. Now the denoising process is conditional on the reference features, as $\mathbf{z}_{l-1} = \mathbf{z}_l - \boldsymbol{\epsilon}_\theta(\mathbf{z}_l, \mathbf{z}^\mathbf{r}, l)$. 

To model the joint distribution, we generalize the latent diffusion process to handle multiple latent codes. 
Specifically, in the forward pass, we use the encoder $\mathcal{E}$ to separately encode the appearance and depth, as $\mathbf{z}^\mathbf{x}_0 = \mathcal{E}(\mathbf{x})$ and $\mathbf{z}^\mathbf{d}_0 = \mathcal{E}(\mathbf{d})$. We then independently add noise to each latent code, as follows:
\begin{align}
    \mathbf{z}^\mathbf{x}_l &= \sqrt{\bar{\alpha}_l} \mathbf{z}^\mathbf{x}_0 + \sqrt{1 - \bar{\alpha}_l} \boldsymbol{\epsilon}^{\mathbf{x}}, \\
    \mathbf{z}^\mathbf{d}_l &= \sqrt{\bar{\alpha}_l} \mathbf{z}^\mathbf{d}_0 + \sqrt{1 - \bar{\alpha}_l} \boldsymbol{\epsilon}^{\mathbf{d}},
\end{align}
where $\boldsymbol{\epsilon}^{\mathbf{x}}$ and $\boldsymbol{\epsilon}^{\mathbf{d}}$ are independently sampled from $\mathcal{N}(0, \mathbf{I})$. This follows the same reverse process as the original latent diffusion model, except now the denoising network $\boldsymbol{\epsilon}_{\theta}$ is modified to denoise both latent codes. For simplicity, we concatenate the noised appearance and depth latent codes, as $\mathbf{z}_l = [\mathbf{z}^\mathbf{x}_l, \mathbf{z}^\mathbf{d}_l]$. Then the denoising network $\boldsymbol{\epsilon}_\theta(\cdot)$ is modified to denoise the concatenated latent code, as $ [\mathbf{z}^\mathbf{x}_{l-1}, \mathbf{z}^\mathbf{d}_{l-1}]  = \mathbf{z}_l - \boldsymbol{\epsilon}_\theta(\mathbf{z}_l,  \mathbf{z}^\mathbf{r}, l)$. During training, $\boldsymbol{\epsilon}_{\theta}$ learns to predict the concatenated noise:
\begin{equation}
    \mathcal{L}(\theta) = \mathbb{E}_{\boldsymbol{\epsilon}^* \sim \mathcal{N}(0, \mathbf{I}), l \sim \mathcal{U}(L)} \left\| [\boldsymbol{\epsilon}^{\mathbf{x}},\boldsymbol{\epsilon}^{\mathbf{d}}] - \boldsymbol{\epsilon}_\theta([\mathbf{z}^\mathbf{x}_l, \mathbf{z}^\mathbf{d}_l],  \mathbf{z}^\mathbf{r}, l) \right\|^2 .
\end{equation}

\subsection{Network Architecture}

\paragraph{Diffusion Backbone.}
We aim to leverage the expressive knowledge stored in a pretrained latent diffusion model to learn our proposed conditional joint distribution with limited available data. 
However, since the latent diffusion model is originally trained to generate only RGB images, it must be adapted to co-generate depth and appearance. Here, we adopt a straightforward solution: expanding the input and output channels of the latent denoising network $\boldsymbol{\epsilon}_\theta$. Specifically, the additional parameters in the input layer are initialized to zero, while the parameters in the output layer are duplicated from the original ones. We find it sufficient for our task, likely due to the rich priors learned in pretrained model, which enhances the model's capability to produce satisfactory results.

\paragraph{ReferenceNet.}
ReferenceNet is designed to enhance and stabilize the generation process by leveraging existing images as reference. It mirrors the layer structure of the denoising model, ensuring compatibility. Both networks produce feature maps with matching spatial resolutions and semantically aligned characteristics. This alignment allows ReferenceNet to effectively integrate extracted features into the diffusion model, resulting in improved visual quality. The weights of our ReferenceNet are initialized from the denoising network and trained together with it.

\begin{figure}
    \includegraphics[width=1.0\linewidth]{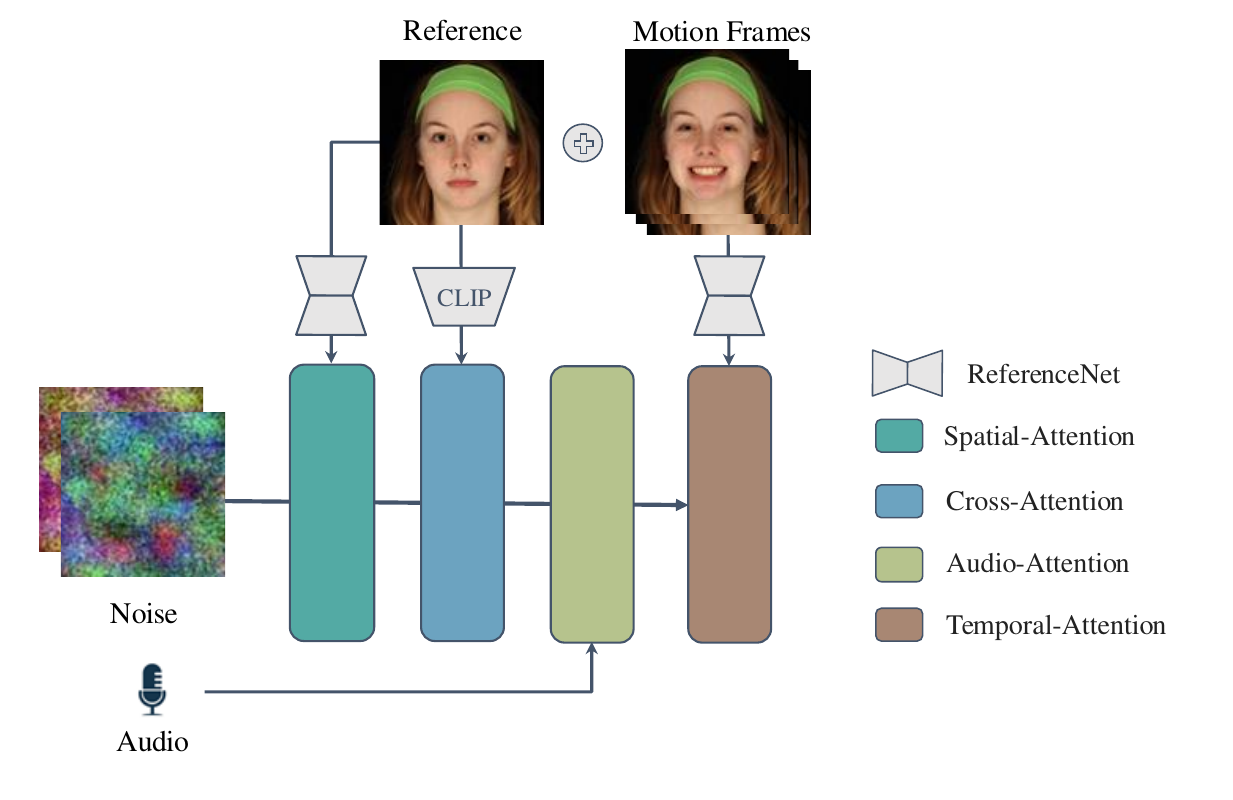}
    \caption{The detailed architecture of the building block of our extended model for portrait RGBD video generation. The model is equipped with additional attention modules to incorporate motion-related inputs.}
    \label{fig:dynamic}
    
\end{figure}

\section{Applications}
Our model can be easily adapted to achieve a wide range of applications. In \secref{sec:bi-directional}, we demonstrate the bi-directional prediction of image or depth conditioned on the other signal.  This allows for tasks such as monocular depth estimation and depth-based image editing. In \secref{sec:video}, we show our model, when equipped with additional motion attention modules, can be extended to generate portrait RGBD videos.

\subsection{Bi-Directional Prediction}
\label{sec:bi-directional}
Our joint distribution of depth and appearance can be transformed into a conditional distribution in both directions, enabling bi-directional prediction. This is accomplished by domain-wise inpainting for our jointly learned model. With a light fine-tuning process, our model is capable of both depth-to-image generation and image-to-depth generation. 
Specifically, we employ masked latent as an additional input condition~\cite{rombach2022high} and design asymmetric masks for appearance and depth while fine-tuning. The task of image-to-depth, \ie monocular depth estimation, can then be achieved by setting pure white for the depth mask and pure black for the image mask, and vice-versa for depth-to-image generation. Note that the involvement of ReferenceNet here enables the generation of the RGB image matching the appearance of the reference image, which allows possibilities for various applications like facial attribute editing and animation. All of these entail the alteration of the depth map, which can be easily achieved with standard 3D editing tools.  We will show both image-to-depth and depth-to-image experiments in \secref{sec:image2depth} and \secref{sec:depth2image}, respectively.


  \begin{figure*}[th]
    \centering
    \includegraphics[width=1.0\linewidth]{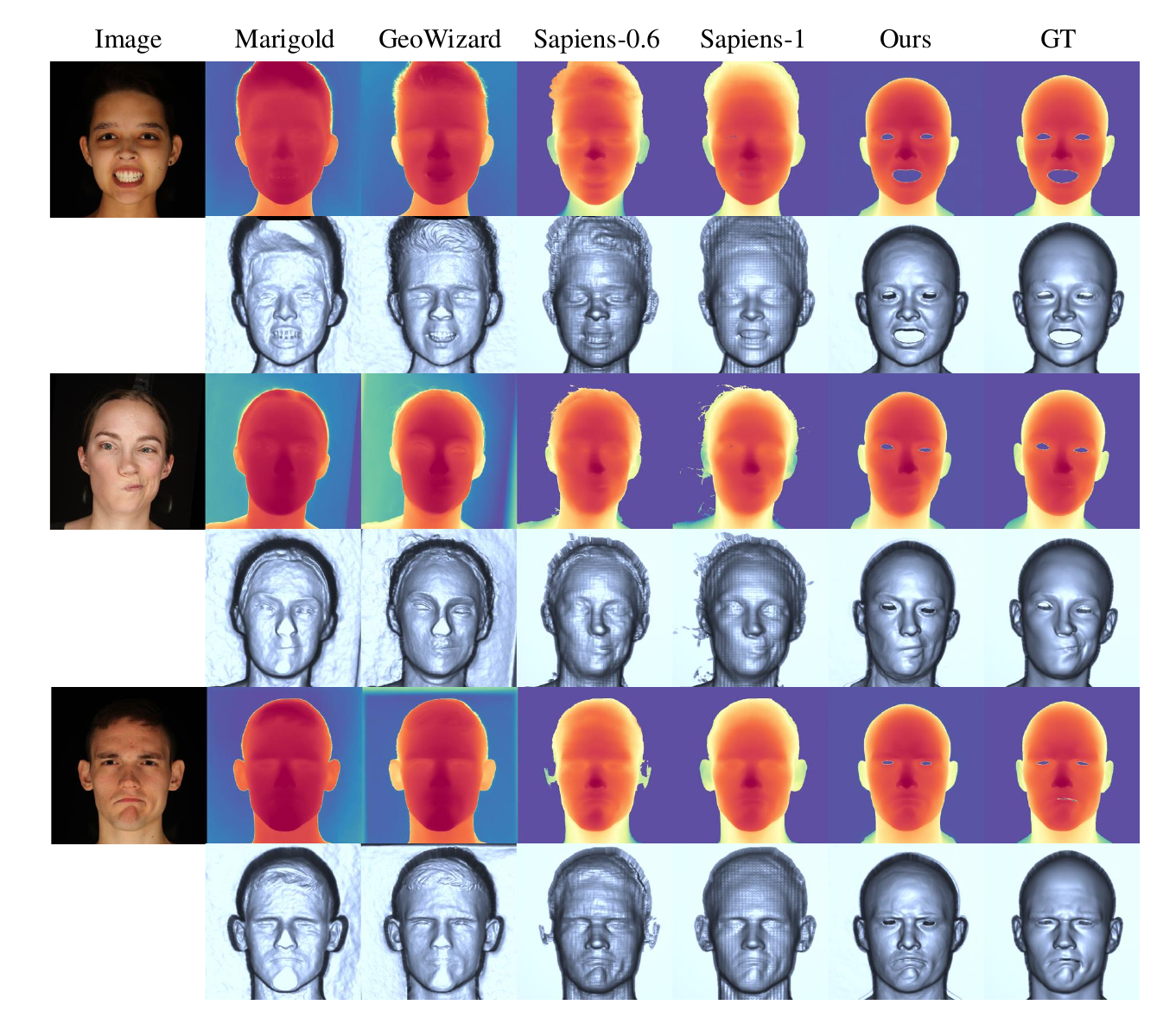}
    \caption{Qualitative comparisons with the state-of-the-art methods for monocular depth estimation on studio images.}
    \label{fig:depth_com_studio}
  \end{figure*}

\subsection{Portrait RGBD Video Generation}
\label{sec:video}
Our model can be easily extended to generate portrait RGBD videos by incorporating motion-related inputs and modules. For example, in the context of audio-driven portrait video generation, audio and temporal feedback are often incorporated into the diffusion backbone with attention modules. Similar to existing methods~\cite{xu2024hallo, tian2024emo, jiang2024loopy}, we add an additional audio-attention module and a temporal-attention module in each building block of the denoising U-Net to extract speech-related motion signals and maintain consistency between the generated frames. \figref{fig:dynamic} illustrates the detailed architecture of the building block of our extended model. Here we use a pretrained Wav2Vec model~\cite{baevski2020wav2vec} to extract a per-frame audio representation and concatenate features of adjacent frames as audio input. To ensure the continuity between consecutive sequences, our model incorporates the last motion frames from previous generated sequence as input to the temporal module. This allows the model to generate consistent portrait RGBD videos that are synchronized with the audio input. Experiments of audio-driven facial animation are illustrated in \secref{sec:audio2face}.

\section{Experiments}
\label{sec:exp}

\begin{figure*}[th]
  \centering
  \includegraphics[width=1.0\linewidth]{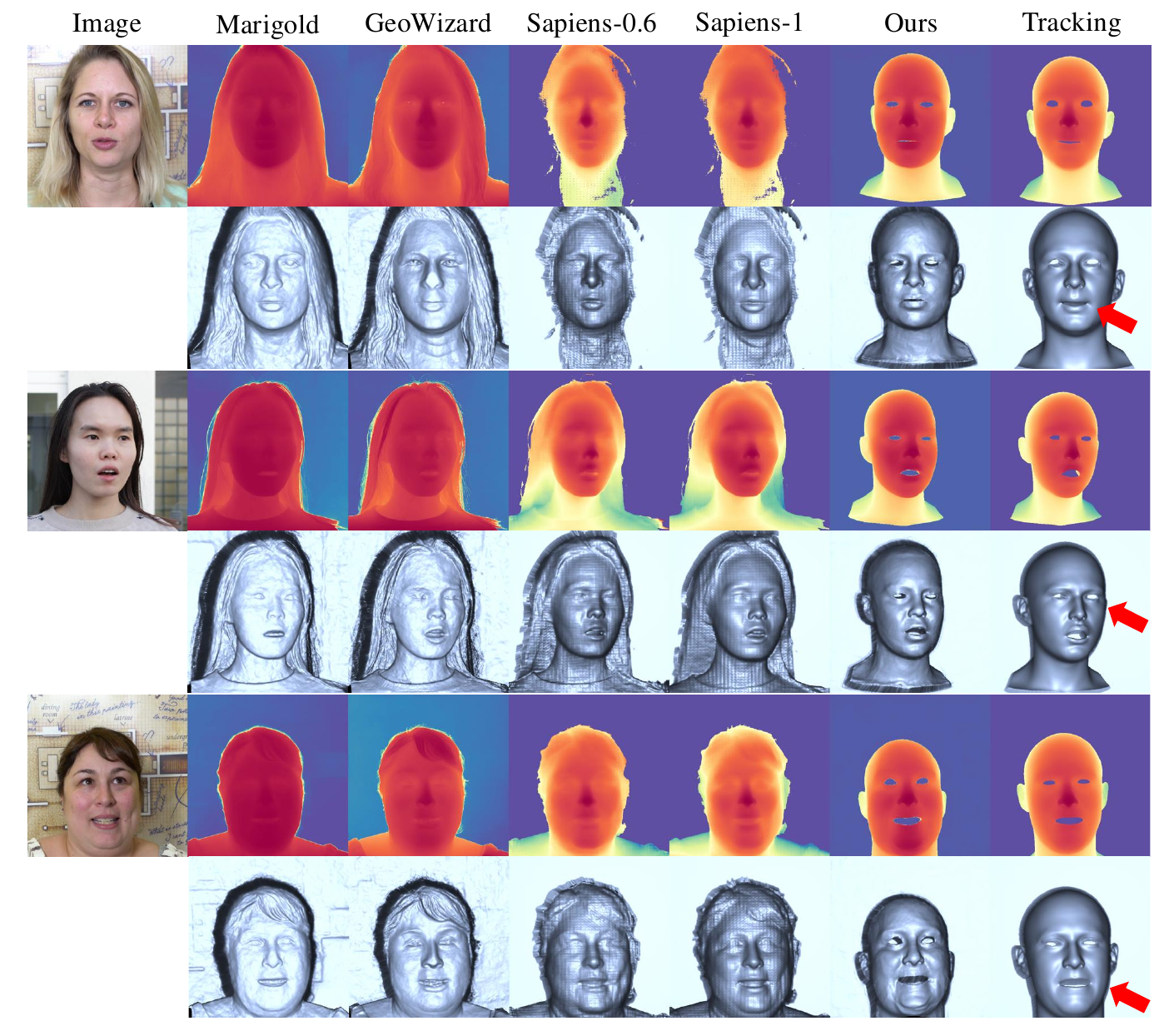}
  \caption{Qualitative comparisons with the state-of-the-art methods for monocular depth estimation on wild faces.  Note that even 3D facial tracking (right column) can sometimes fail.  Our method can achieve a better depth due to the high-quality studio data as a subset of our training data.}
  \label{fig:depth_com_wild}
\end{figure*}

\subsection{Experimental Setups}
\label{sec:exp_setup}
\noindent {\bf Implementation Details.}
The joint learning of portrait RGB images and depth takes about three days on four 4090 GPUs. We use a batch size of 32 and a constant learning rate of 1e-5 and train our model for 30000 steps. To learn our extended model with audio conditioning, we use a batch size of 1. The weights of the motion module are initialized from Animatediff\cite{guo2023animatediff}, and we retain all other parameters from the first stage. We generate 14 frames at once and use the first 4 ground truth frames of each training sample as the motion frames during training.

\noindent {\bf Datasets.}
We train our model on a combination of datasets collected from both studio and in-the-wild scenarios. Particularly, we use a high-quality multi-view studio face dataset~\cite{chandran2020semantic}, comprising of images from 336 subjects performing various facial expressions, along with corresponding high-fidelity registered meshes. We render ground truth depth maps from these registered meshes to obtain the paired RGB-Depth data for training our method. 
To improve the generalization of our model to real world data, we incorporate in-the-wild audio-visual sequences, from selected clips of the HDTF~\cite{zhang2021flow} and VFHQ~\cite{xie2022vfhq} datasets. As these in-the-wild datasets do not contain corresponding geometry, we use a state of the art monocular face tracking approach~\cite{chandran2023continuous, chandran2024infinite} to estimate 3D geometry from these videos, using which we can extract pseudo ground truth depth maps for training. In total we collect around 3 hours of video data from HDTF and VFHQ, containing 2,423 clips with diverse identities, facial expressions and head poses. All videos are sampled at 25 FPS and the images are cropped to a resolution of \(512\times512\). We also pre-process the audio to 16kHz and extract per-frame wav2vec audio features. The combination of studio data and in-the-wild data provides a solid foundation, enabling our network to jointly generate high-quality image and depth across various practical scenarios (see \cref{fig:depth_com_studio,fig:depth_com_wild}). We will now demonstrate and evaluate the many different applications made possible by our approach.


\begin{figure}[th]
  \centering
  \includegraphics[width=1.0\linewidth]{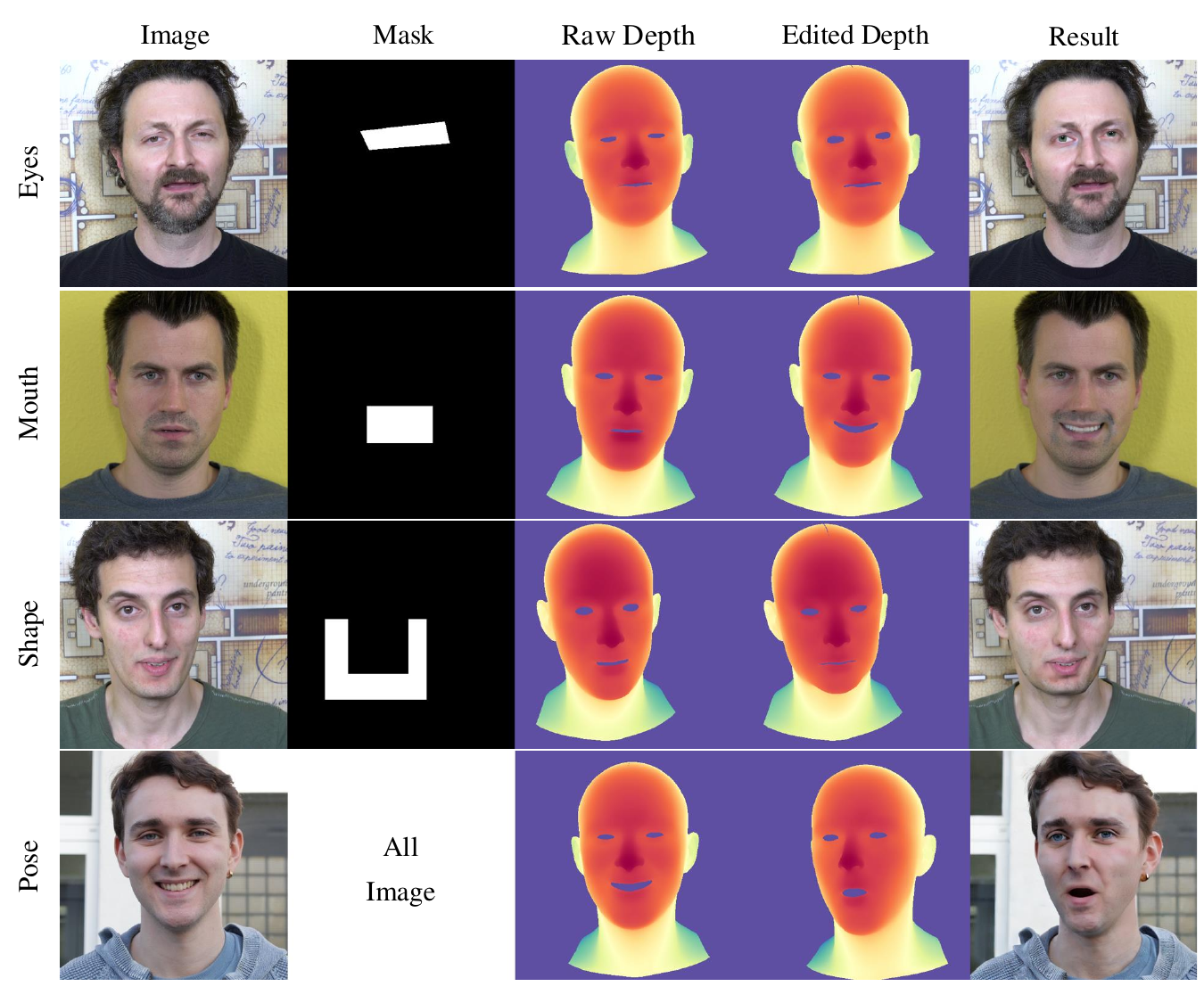}
  \caption{Depth-based face editing on shape, expression and pose.}
  \label{fig:editing}
\end{figure}

\subsection{Depth Estimation}
\label{sec:image2depth}

Recently there has been great interest in fine-tuning foundational models to predict depth from monocular RGB input~\cite{fu2025geowizard, yang2024depth, ke2024repurposing, guizilini2023towards}. As illustrated in \secref{sec:bi-directional}, our model can readily be used for the task of monocular depth estimation (or RGB conditioned depth prediction) after a light fine-tuning. 

We evaluate depth estimation on an unseen studio dataset as it provides us with precise 3D depth maps which we can consider as ground truth. This evaluation dataset consists of 1264 images from 55 identities consisting of various facial expressions and poses. We follow the relative-depth evaluation protocols proposed in MiDaS~\cite{birkl2023midas} and LDM3D~\cite{stan2023ldm3d}, and evaluate standard metrics including absolute relative error (AbsRel), \(\delta_{1}\) accuracy and root mean squared error (RMSE). As the ground truth depth maps derived from the 3D mesh in the studio dataset contain only the facial skin region, we apply a mask and remove regions outside this area to ensure a fair comparison of methods.

We compare our method against state-of-the-art monocular depth estimators including Marigold~\cite{ke2024repurposing}, Geowizard~\cite{fu2025geowizard}, and three different backbones from the human-centric foundation model Sapiens~\cite{khirodkar2025sapiens}. Quantitative results are listed in \cref{tab:depth} and qualitative results are shown in \cref{fig:depth_com_studio}. As we see in \cref{tab:depth}, our method outperforms all other models on this task, including the Sapiens-1B model. Qualitatively our method captures the facial shape and expression similar to Sapiens-1B, while containing significantly fewer grid-like artifacts. We also present qualitative results for monocular depth estimation on in-the-wild face portraits (\figref{fig:depth_com_wild}), and compare our estimated depth to the result of fitting a 3D morphable model~\cite{chandran2023continuous, chandran2024infinite} to the input RGB image. Even though such a fitting method was used to generate the depth component of our in-the-wild training data, our results on unseen in-the-wild images have better mouth and face structure when compared to the 3DMM fit, and correspond better to the RGB image.  This is due to the fact that our method learned accurate depth correlation from the combined studio training data.


\begin{table}
    \centering
    \begin{tabular}{l| c c c}
      \toprule
      Method & AbsRel $\downarrow$ & $\delta_{1}$ $\uparrow$ & RMSE $\downarrow$ \\
      \midrule
      Marigold~\cite{ke2024repurposing} & 0.529 & 0.538 & 0.055 \\
      GeoWizard~\cite{fu2025geowizard} & 0.392 & 0.644 & 0.050\\
      Sapiens-0.3B~\cite{khirodkar2025sapiens} & 0.313 & 0.526 & 0.056\\
      Sapiens-0.6B~\cite{khirodkar2025sapiens} & 0.297 & 0.549 & 0.048\\
      Sapiens-1B~\cite{khirodkar2025sapiens} & 0.197 & 0.696 & \textbf{0.047}\\
      \midrule
      Ours-Wild-Only & 0.313  & 0.658 & 0.059   \\
      Ours & \textbf{0.162}  & \textbf{0.765} & \textbf{0.047}   \\
      \bottomrule
    \end{tabular}
    \caption{Quantitative comparison for monocular depth estimation on portrait images.}
    \label{tab:depth}
    \vspace{-1mm}
  \end{table}
\begin{figure}[th]
  \centering
  \includegraphics[width=1.0\linewidth]{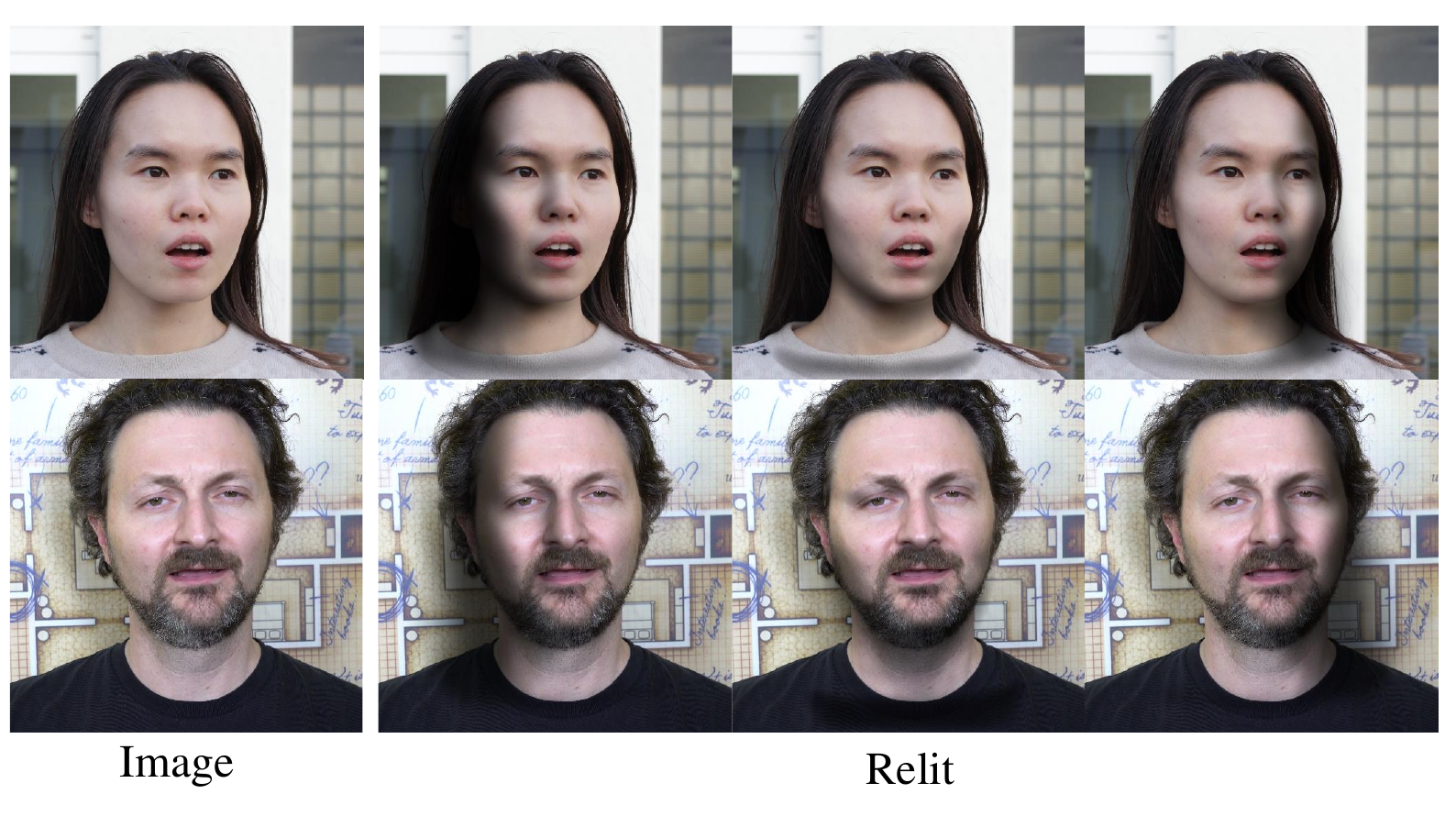}
  \caption{Portrait image relighting is possible using our generated depth map.}
  \label{fig:relighting}
  \vspace{-2mm}
\end{figure}

  \subsection{Depth-Conditioned Portrait Animation}
\label{sec:depth2image}

  In addition to monocular depth estimation, the joint learning of RGB and depth modalities also enables us to generate an RGB image by providing a depth map as input. This application of our model can be particularly useful in having precise control over the generated RGB image for editing applications. In \cref{fig:editing}, we show examples of editing an RGB image, by modifying its corresponding depth map, and requiring our model to re-generate an RGB image corresponding to the edited depth. The inpainting mask spatially guides the model to the regions it is expected to modify in the given image. Our approach generates photo real images that respect the identity of the original RGB image and the edited depth maps.


  \subsection{Image Relighting}
  One benefit of our joint learning of appearance and depth is that the facial depth can be used for downstream tasks like portrait relighting. \figref{fig:relighting} illustrates an example where the generated depth maps are used to compute surface normals for basic lighting changes in the generated image.  Here the normals are used for rendering a diffuse shading layer that is multiplied with the image as a post-process.

\subsection{Audio-driven Facial Animation}
\label{sec:audio2face}
\begin{figure}[t]
  \centering
  \includegraphics[width=1.0\linewidth]{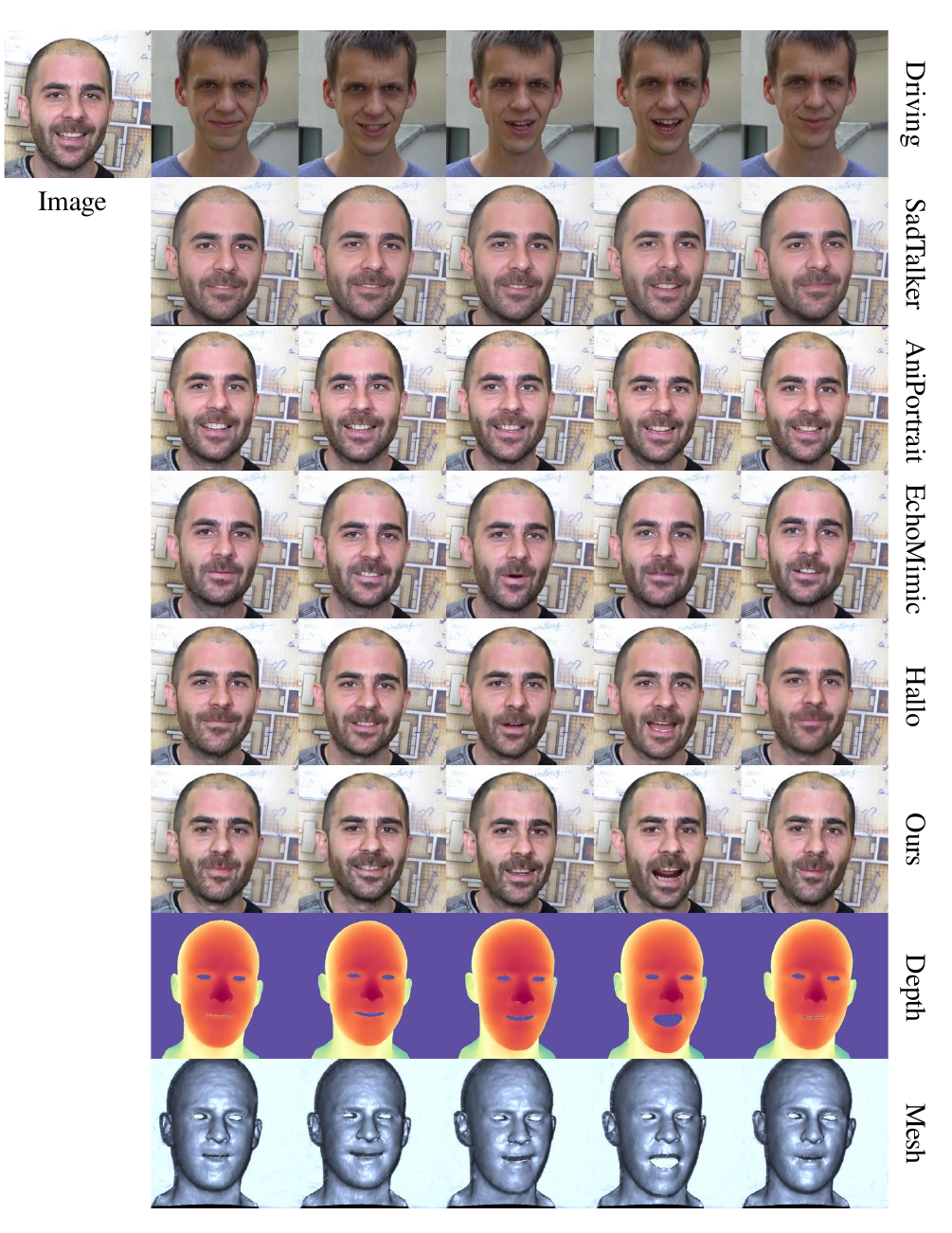}
  \caption{Qualitative comparison with existing audio-driven animation methods. Note that our method can jointly generate image and depth animation.}
  \vspace{-3mm}
  \label{fig:audio}
\end{figure}

So far the applications we have seen in monocular depth estimation and depth-conditioned portrait editing focus on single-image manipulation.  However as we described in \cref{sec:video}, our network can be extended with the insertion of motion modules and also be used for audio driven facial animation, when trained along with in-the-wild audio visual data. In \cref{fig:audio}, we compare the results of using our method for audio driven facial animation against other 2D audio-driven methods including SadTalker~\cite{zhang2023sadtalker}, AniPortrait~\cite{wei2024aniportrait}, EchoMimic~\cite{chen2024echomimic} and Hallo~\cite{xu2024hallo}. While providing visually similar results for the RGB channels, our method can also jointly generate depth maps that align with the generated images, thereby introducing new capability that is missing in existing audio driven facial animation techniques.

\subsection{Ablation Studies}
\begin{figure}[t]
  \centering
  \includegraphics[width=1.0\linewidth]{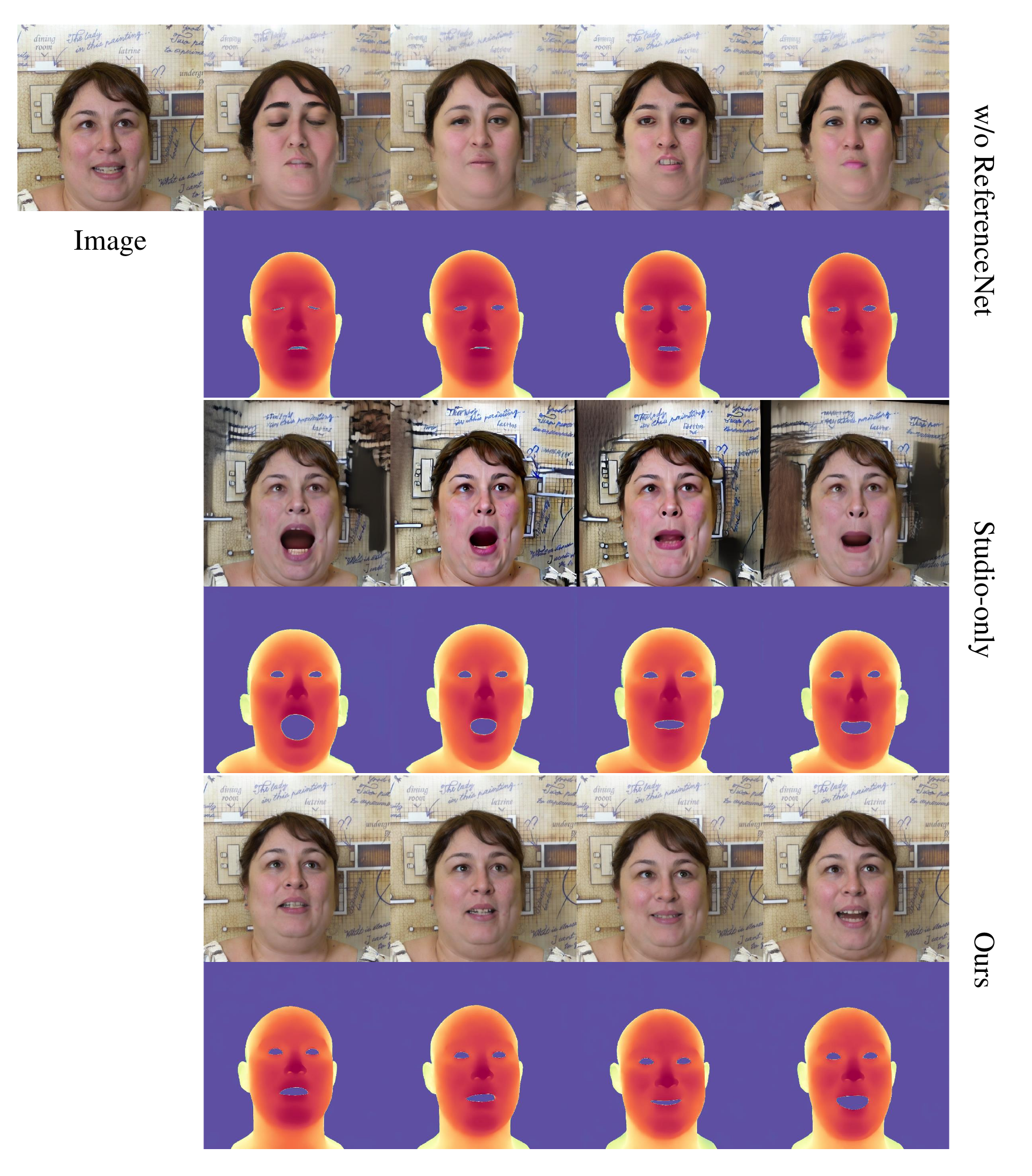}
  \vspace{-5mm}
  \caption{Ablation studies to show the effect of our architecture without the ReferenceNet (top), and trained on studio-only data (middle), compared to our proposed method (bottom).}
  \label{fig:ablation}
  \vspace{-3mm}
\end{figure}

We first evaluate the influence of the ReferenceNet on the quality of our generated results. We train a version of our network where we remove the ReferenceNet, and instead provide the latent reference RGB image as an additional input to the denoising U-Net. After training, we jointly generate RGB images and depth maps from multiple different noise inputs, which are shown in the first row of \cref{fig:ablation}. The generations without the ReferenceNet fail to capture the identity of the reference image, highlighting its importance in our architecture.

Secondly we also evaluate the importance of training our method on both studio data with ground truth depth, and in-the-wild data with pseudo ground truth depth. We first verify whether our method trained only on studio data can generalize to unseen in-the-wild identities. As we seen in the second row of \cref{fig:ablation}, training only on studio data results in poor generalization to in-the-wild data and degrades the visual quality of the generated RGB images. However training with data from both studio and in-the-wild sources, results in the best performance as we see in the last row of the \cref{fig:ablation}. This is also confirmed by our quantitative results in \cref{tab:depth}.


\subsection{Limitations}
Although our method is not limited to the facial skin region in principle, as it currently relies on depth maps derived from registered 3D geometry for training, it can only predict depth maps only for the skin region when given a new RGB image. Secondly, due to our modest computational resources, we were limited from scaling our training datasets and training times to those comparable with existing audio driven portrait animation methods. Therefore our current results could also be substantially improved with longer training on larger datasets.
\section{Conclusion}
\label{sec:conclusion}

In this work we propose a new generative model for face portrait images, with a focus on jointly learning the visual appearance and the 3D depth in a unified framework.  To accomplish this task we introduce a new diffusion-based architecture and corresponding training scheme, which ensures correlation between the two different output signals.  After training, our model can be used in several image manipulation and animation applications.  Here we have demonstrated tasks such as coupled image+depth portrait generation, monocular facial depth estimation, depth-based image editing, portrait image relighting and audio-driven talking head synthesis.  Further applications are also possible with our joint learning framework, which believe advances the state-of-the-art in generative portrait image modeling.
{
    \small
    \bibliographystyle{ieeenat_fullname}
    \bibliography{main}
}

\clearpage
\setcounter{page}{1}
\maketitlesupplementary

In this supplementary material, we provide more details about the network architecture and data processing. More information on our training details and more results of our method are also provided. We strongly recommend watching the supplementary video.  

\section{Implementation Details}
\label{Implementation}
We describe more implementation details on the network architecture and the training details used in \secref{sec:method} of the main paper.

\subsection{Network Architecture}
\label{sec:s1}
\begin{figure}
    \includegraphics[width=1.0\linewidth]{{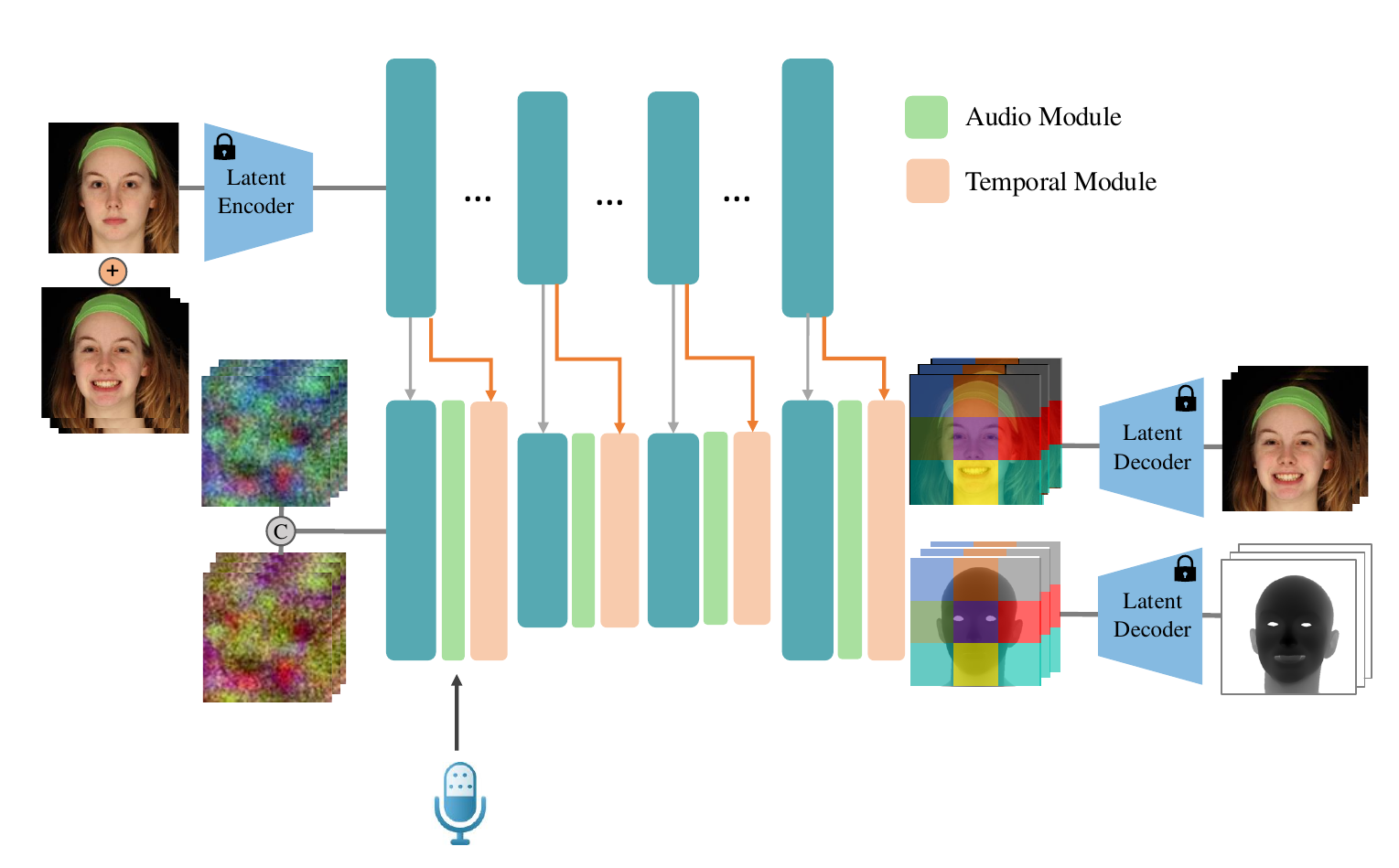}}
    \caption{The overview of our audio-driven animation network.}
    \label{fig:supp_pipe}
\end{figure}

Here, we present the details of our network architecture for audio-driven animation (\cref{sec:video}), as shown in \cref{fig:supp_pipe}. For the audio-driven animation task, we add an audio-attention module and a temporal-attention module in each block.

\begin{itemize}
    \item \textbf{Audio-Attention Module.} 
    Our method effectively adopts a large pre-trained speech model wav2vec~\cite{baevski2020wav2vec} for the downstream task of audio-driven facial animation generation. 
    Here we concatenate the audio features extracted by different blocks of the wav2vec model to get an audio representation for each frame. After getting the per-frame audio feature, we use the feature of a window of k continuous frames as the input of the center frame.
    Then we perform cross-attention between this audio feature and the input to the audio module, extracting motion signals from the speech.

    \item \textbf{Temporal-Attention Module.} 
    We use the same Temporal-Attention layers as in recent advances~\cite{ tian2024emo}. This module is designed to ensure smooth transitions across synthesized frames. To capture the dependencies between consecutive frames, we apply self-attention mechanisms on the temporal dimension of the features.
    Specifically, we first reshape the input feature $F \in \mathbb{R}^{ b \times c \times f \times h \times w} $, where $ b, c, f, h, w$ represent the batch size, feature channel, the number of the generated frames in a sequence and the height and width of the feature map, to  $F \in \mathbb{R}^{ (b \times h \times w )\times c \times f} $. Then we apply self-attention across the temporal dimension $f$. However, motion consistency can only be guaranteed inside each sequence in this way, constraining the application for long video generation. Therefore, we draw inspiration from existing works~\cite{wang2021audio2head} and take the last $ n$ generated frames from the preceding sequence as the motion frames. Here, we first feed these motion frames into the ReferenceNet to extract multi-resolution motion features. Then in each block, we concatenate the temporal module input and the motion feature along the temporal dimension $f$ to get the self-attention layer input. In this way, the motion information from the previous sequence can be involved, so to ensure the coherence among different clips.

\end{itemize}

\begin{figure*}[th]
    \centering
    \includegraphics[width=1.0\linewidth]{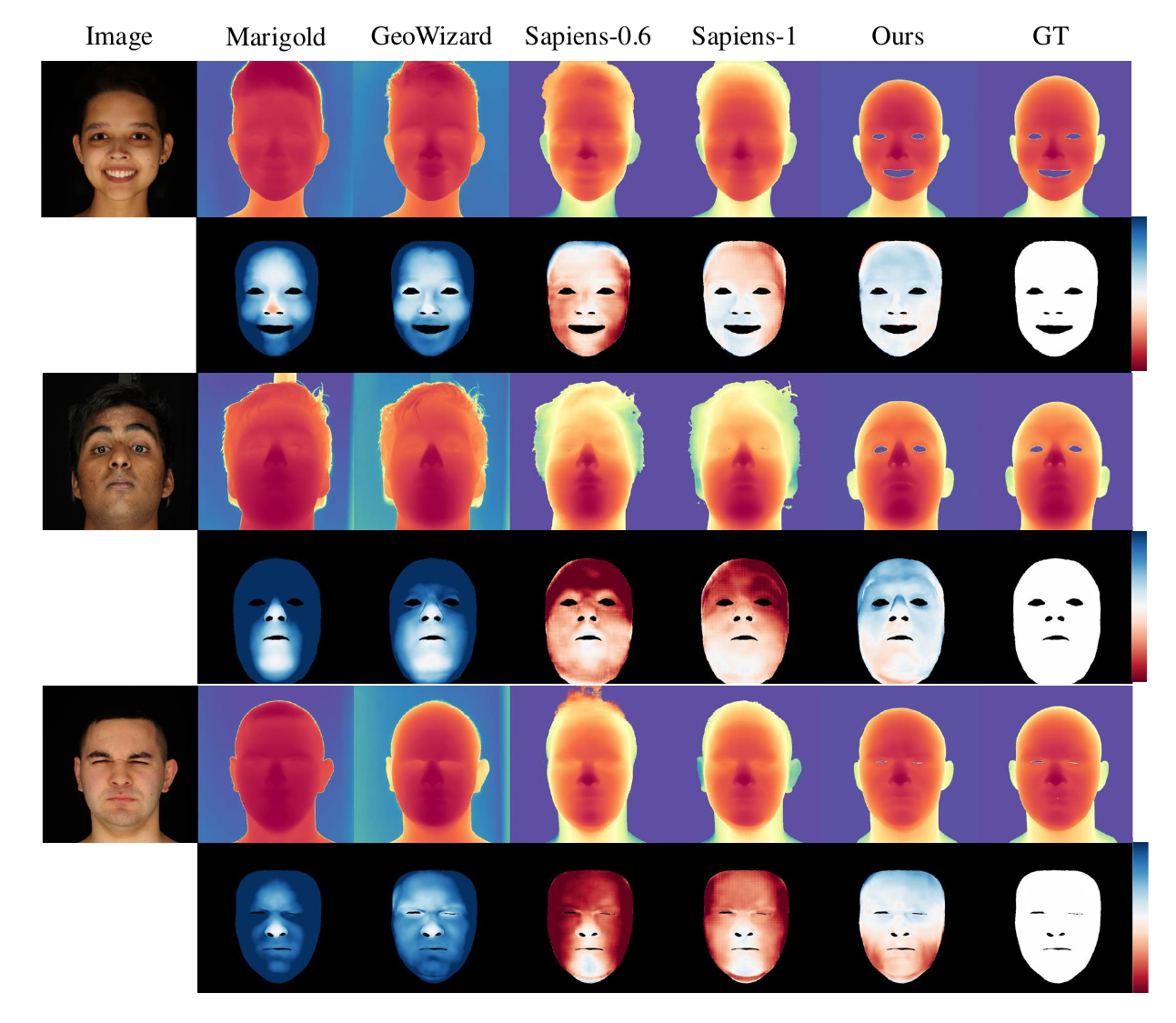}
    \caption{Qualitative comparisons with the state-of-the-art methods for monocular depth estimation on studio faces. We also show error map under facial mask for each sample. Note that here white means no error.}
    \label{fig:error_map}
  \end{figure*}

\subsection{Training Details}

As described in \cref{sec:method}, our method contains three different training stages. In the first stage, we use the multi-view studio face dataset along with the in-the-wild dataset to train our joint-learning network. In the second stage, we slightly fine tune our model by incorporating different masks for the inpainting task. Here we design asymmetric masks for the RGB and depth branches, so that after fine-tuning, our model is capable of down-stream tasks including depth estimation, relighting and depth-based image editing. Note that our model is still capable of jointly generating RGB and depth by setting the mask to be pure white for both domains. In the third stage, we extend our first stage model to the audio-driven RGBD animation task by incorporating audio and temporal attention modules. Only in-the-wild datasets are employed in this stage due to the lack of audio and video data in studio face datasets. As mentioned in \cref{sec:exp_setup}, we fix the parameters from the first stage while training. The training is performed on 4x RTX 4090 GPUs, and it takes about 3 days for the joint-learning model training, 15 hours for the inpainting fine-tuning and 2 days for the motion module training respectively.

\subsection{Data Processing}

In order to obtain the corresponding depth map for monoculor in-the-wild datasets, we use an off-the-shelf face tracking tool~\cite{chandran2023continuous, chandran2024infinite} to fit a face mesh for each frame and then render out the depth map. The fitted mesh is represented by the blendshape weights of a PCA-based face model, which includes 50 eigen faces for identity and 25 for expression. A landmark loss and a photometric loss are utilized to optimize the weights. To ensure the stability and smoothness of the tracking on a video sequence, we solve a global identity code for each clip.

\section{More Results}

\subsection{Depth Estimation}

As demonstrated in \cref{sec:image2depth}, we apply a mask on the face skin region while calculating the depth estimation metrics with ground truth. In \cref{fig:error_map}, we show more qualitative results along with the error map under the mask. Here we set the error range to be -0.1 to 0.1. As demonstrated in the figure, our method outperforms other methods for generating depth maps with accurate geometry under various expressions and poses.  



\end{document}